\documentclass{article}

\usepackage{microtype}
\usepackage{graphicx}
\usepackage{booktabs} 
\usepackage{amsmath, amssymb, bbm, mathtools}
\usepackage[caption=false, format=hang]{subfig}

\usepackage{hyperref}

\newcommand{\RR}{\mathbb{R}}
\newcommand{\minimize}[2]{\ensuremath{\underset{\substack{{#1}}}{\mathrm{minimize}}\;\;#2 }}

\usepackage[normalem]{ulem} 

\usepackage[draft]{icml2020}

\begin{document}

\twocolumn[
\icmltitle{Wasserstein-based Graph Alignment}



\icmlsetsymbol{equal}{*}

\begin{icmlauthorlist}
\icmlauthor{Hermina Petric Maretic}{epfl,equal}
\icmlauthor{Mireille El Gheche}{epfl,equal}
\icmlauthor{Matthias Minder}{epfl}
\icmlauthor{Giovanni Chierchia}{esiee}
\icmlauthor{Pascal Frossard}{epfl}
\end{icmlauthorlist}

\icmlaffiliation{epfl}{Ecole Polytechnique F\'ed\'erale de Lausanne, Signal Processing Laboratory (LTS4), Lausanne, Switzerland}
\icmlaffiliation{esiee}{Laboratoire d'informatique Gaspard-Monge, CNRS, ESIEE Paris, Univ Gustave Eiffel, Marne-la-Vall\'ee, France}

\icmlcorrespondingauthor{Pascal Frossard}{pascal.frossard@epfl.ch}

\icmlkeywords{Machine Learning, Graphs, Optimal Transport, Optimization, Mapping}

\vskip 0.3in
]



\printAffiliationsAndNotice{\icmlEqualContribution} 


\begin{abstract}
We propose a novel method for comparing non-aligned graphs of different sizes, based on the Wasserstein distance between graph signal distributions induced by the respective graph Laplacian matrices. Specifically, we cast a new formulation for the one-to-many graph alignment problem, which aims at matching a node in the smaller graph with one or more nodes in the larger graph. By integrating optimal transport in our graph comparison framework, we generate both a structurally-meaningful graph distance, and a signal transportation plan that models the structure of graph data. The resulting alignment problem is solved with stochastic gradient descent, where we use a novel Dykstra operator to ensure that the solution is a one-to-many (soft) assignment matrix. We demonstrate the performance of our novel framework on graph alignment and graph classification, and we show that our method leads to significant improvements with respect to the state-of-the-art algorithms for each of these tasks.
\end{abstract}

\section{Introduction}
\label{introduction}

The importance of graphs has recently increased in various tasks in different application domains, such as molecules modeling, brain connectivity analysis, or social network inference. Even if this development is partially fostered by powerful mathematical tools to model structural data, important questions are still largely open. In particular, it remains challenging to align, classify, predict or cluster graphs, since the notion of similarity between graphs is not straightforward. In many cases (e.g., dynamically changing graphs, multilayer graphs, etc\dots), even a consistent enumeration of the vertices cannot be trivially chosen for all graphs under study.

When two graphs are not aligned \textit{a priori}, graph matching must be performed prior to any comparison, leading to the challenging problem of estimating an unknown assignment between their vertices. Since this problem is NP-hard, there exist several relaxations that can be solved by minimizing a suitable distance between graphs under the quadratic assignment model, such as the $\ell_2$-norm between the graph adjacency matrices \cite{NIPS2018_7365}, or the Gromow-Wasserstein distance \cite{NIPS2019_8569}. However, these approaches may yield solutions that are unable to capture the importance of edges with respect to the overall structure of the graph. 
An alternative that seems more appropriate for graph comparison is based on the Wasserstein distance between the graph signal distributions \cite{NIPS2019_9539}, but it is currently limited to graphs of the same size. 

In this paper, we consider the challenging alignment problem for graphs of different sizes. In particular, we build on \cite{NIPS2019_9539} and formulate graph matching as a one-to-many soft-assignment problem, where we consider the Wasserstein distance to measure the goodness of graph alignment in a structurally meaningful way. To accommodate for the nonconvexity of the problem, we propose a stochastic formulation based on a novel Dykstra operator to implicitly ensure that the solution is a one-to-many soft-assignment matrix. This allows us to devise an efficient algorithm based on stochastic gradient descent, which naturally integrates Bayesian exploration in the optimization process, so as to help finding better local minima. We illustrate the benefits of our new graph comparison framework in representative tasks such as graph alignment and graph classification on synthetic and real datasets. Our results show that the Wasserstein distance combined with the one-to-many graph assignment permits to outperform both Gromov-Wasserstein and Euclidean distance in these tasks, suggesting that our approach outputs a structurally meaningful distance to efficiently align and compare graphs. These are important elements in graph analysis, comparison, or graph signal prediction tasks.

The paper is structured as follows. Section \ref{sec:problem_formulation} presents the graph alignment problem with optimal transport, as well as the formulation of the one-to-many assignment problem. Section \ref{sec:optimization} introduces our new Dykstra operator and proposes an algorithm for solving the resulting optimization problem via stochastic gradient descent. In Section \ref{sec:experiments}, the performance of the proposed approach is assessed on synthetic and real data, and compared to different state-of-the-art methods. Finally, Section \ref{sec:conclusion} concludes the paper.

\section{Related work}
\label{sec:related_work}
Numerous methods have been developed for graph alignment, whose goal is to match the vertices of two graphs such that the similarity of the resulting aligned graphs is maximized. This problem is typically formulated under the quadratic assignment model \cite{YanICMR2016,NIPS2017_6911}, which is generally thought to be essential for obtaining a good matching, despite being NP-hard. The main body of research in graph matching is thus focused on devising more accurate and/or faster algorithms to solve this problem approximately \cite{beamsearch_2006}.

In order to deal with the NP-hardness of graph alignment, spectral clustering based approaches \citep{Caelli2004, Srinivasan2006} relax permutation matrices into semi-orthogonal ones, at the price of a suboptimal matching accuracy. Alternatively, semi-definite programming can be used to relax the permutation matrices into semi-definite ones \citep{Schellewald2005}. Spectral properties have also been used to inspect graphs and define different classes of graphs for which convex relaxations are tight \cite{aflalo2015convex, fiori2015spectral, dym2017ds++}. Based on the assumption that the space of doubly-stochastic matrices is a convex hull of the set of permutation matrices, the graph matching problem was relaxed into a nonconvex quadratic problem \cite{cho2010reweighted, Zhou2012}. A related approach was recently proposed to approximate discrete graph matching in the continuous domain by using nonseparable functions \cite{NIPS2018_7365}. Along similar lines, a Gumbel-sinkhorn network was proposed to infer permutations from data \cite{Mena2018, Emami2018} and align graphs with the Sinkhorn operator \cite{Sinkhorn1964} to predict a soft permutation matrix. 

Closer to our framework, some recent works studied the graph alignment problem from an optimal transport perspective. Flamary \emph{et al.} \cite{flamary:hal-01103076} proposed a method to compute an optimal transportation plan by controlling the displacement of vertex pairs. Gu \emph{et al.} \cite{GU201556} defined a spectral distance by assigning a probability measure to the nodes via the spectrum representation of each graph, and by using Wasserstein distances between probability measures. This approach however does not take into account the full graph structure in the alignment problem. Later, Nikolentzos \emph{et al.} \cite{nikolentzos2017matching} proposed instead to use the Wasserstein distance for matching the graph embeddings represented as bags of vectors.

Another line of works looked at more specific graphs. Memoli \cite{memoli2011gromov} investigated the Gromov-Wasserstein distance for object matching, Peyré \emph{et al.} \cite{2016-peyre-icml} proposed an efficient algorithm to compute the Gromov-Wasserstein distance and the barycenter of pairwise dissimilarity matrices, and \cite{NIPS2019_8569} devised a scalable version of Gromov-Wasserstein distance for graph matching and classification. More recently, Vayer \emph{et al.} \cite{vayer2018optimal} built on this work to propose a distance for graphs and signals living on them, which is a combination between the Gromov-Wasserstein of graph distance matrices, and the Wasserstein distance of graph signals. However, while the above methods solve the alignment problem using optimal transport, the simple distances between aligned graphs do not take into account its global structure and the methods do not consider the transportation of signals between graphs.

\section{Problem Formulation}
\label{sec:problem_formulation}

Despite recent advances in the analysis of graph data, it stays challenging to define a meaningful distance between graphs. Even more, a major difficulty with graph representations is the lack of node alignment, which is necessary for direct quantitative comparisons between graphs. We propose to use the Wasserstein distance to compare graphs \cite{NIPS2019_9539}, since it has been shown to take into account global structural differences between graphs. Then, we formulate graph alignment as the problem of finding the assignment matrix that minimizes the distance between graphs of different sizes. 
 
\subsection{Preliminaries}

\paragraph{Optimal transport} 
Let $(\nu, \mu)$ be the set of two arbitrary probability measures on two spaces $(\mathcal{X}, \mathcal{Y})$. The Wasserstein distance\footnote{Wasserstein distance is also referred to as Kantorovich-Monge-Rubinstein distance.} $\mathcal{W}_2(\nu, \mu)$, arising from the Monge and Kantorovich optimal transport problem, can be defined as finding a map $T: \mathcal{X}   \rightarrow \mathcal{Y}$ that minimizes
\begin{equation}
\label{eq:wass}
\mathcal{W}_2(\nu, \mu) = \inf_{T_{\#} \nu= \mu}{\int_{\mathcal{X}}  \|x - T(x)\|^2 \, d\nu (x)},
\end{equation} 
where $T_{\#} \nu= \mu$ means that $T$ pushes forward the mass from $\nu$ to $\mu$. Intuitively, $T$ can be seen as a function that preserves positivity and total mass, i.e., moving an entire probability mass on $\mathcal{X}$ to an entire probability mass on $\mathcal{Y}$. Equation \eqref{eq:wass} can be seen as the minimal cost needed to transport one probability measure to another with respect to a quadratic cost $c(x,y) = \|x-y\|^2_2$. 

The Wasserstein distance between Gaussian distributions has an explicit expression in terms of their mean vectors and covariance matrices $\Sigma_1$ and $\Sigma_2$, respectively. With $\nu = \mathcal{N}(0, \Sigma_1)$ and $\mu = \mathcal{N}(0, \Sigma_2)$, the above distance can be written as \cite{takatsu2011wasserstein}
\begin{equation}\label{eq:wass_explicit}
\mathcal{W}_2^2\big(\nu, \mu \big) = 
{\rm Tr}\left(\Sigma_1 + \Sigma_2\right) - 2 \, {\rm Tr}\left(\sqrt{\Sigma_1^{\frac{1}{2}} \Sigma_2 \Sigma_1^{\frac{1}{2}}}\right),
\end{equation}
and the optimal map $T$ that takes $\nu$ to $\mu$ is 
\begin{equation}
T(x) = \Sigma_1^{\frac{1}{2}}\Big(\Sigma_1^{\frac{1}{2}} \Sigma_2  \Sigma_1^{\frac{1}{2}}\Big)^{\frac{1}{2}}\Sigma_1^{\frac{1}{2}} x.
\end{equation}

\paragraph{Smooth graph signals} 
Let $\mathcal{G}=(V, E)$ be a graph defined on a set of $N$ vertices, with (non-negative) similarity edge weights. We denote by $W\in\RR^{N \times N}$ the weighted adjacency matrix of $\mathcal{G}$, and $D={\rm{diag}}(d_1, \cdots, d_N)$ the diagonal matrix of vertex degree $d_i=\sum_j w_{ij}$ for all $i$. 
The Laplacian matrix of $\mathcal{G}$ is thus defined as $L = D-W$.

We further assume that each vertex of the graph $\mathcal{G}$ is associated with a scalar feature, forming a graph signal. We denote this graph signal as a vector $x\in \mathbb{R}^{N}$. Following \citep{rue2005gaussian}, we interpret graphs as key elements that drive the probability distributions of signals, and thus we consider that a graph signal follows a normal distribution with zero mean and covariance matrix $L^\dagger$
\begin{equation}
x \sim \nu^{\mathcal{G}} = \mathcal{N}(0, L^\dagger),
\end{equation} 
where $\dagger$ denotes a pseudoinverse operator.
The above formulation means that the graph signal varies slowly between strongly connected nodes \citep{Dong14}. This assumption is verified for most common graph and network datasets. It is further used in many graph inference algorithms that implicitly represent a graph through its smooth signals \citep{dempster1972covariance, friedman2008sparse, dong2018learning}. Furthermore, the smoothness assumption is used as regularization in many graph applications, such as robust principal component analysis \citep{shahid2015robust} and label propagation \citep{zhu2003semi}.

\subsection{One-to-many assignment problem}
Assume that we are given two graphs $\mathcal{G}_1$ and $\mathcal{G}_2$ with the same number of nodes, and that we have knowledge of the one-to-one mapping between their vertices. 

Following \cite{NIPS2019_9539}, instead of comparing graphs directly, we look at their signal distributions, which are governed by the graphs. Specifically, we measure the dissimilarity between two aligned graphs $\mathcal{G}_1$ and $\mathcal{G}_2$ through the Wasserstein distance of the respective distributions $\nu^{\mathcal{G}_1}= \mathcal{N}(0, L_1^\dagger)$ and $\mu^{\mathcal{G}_2}= \mathcal{N}(0, L_2^\dagger)$, which can be calculated explicitly as
\begin{equation}\label{eq:wass_explicit_graph}
\mathcal{W}_2^2\big(\nu^{\mathcal{G}_1}, \mu^{\mathcal{G}_2}\big) = 
{\rm Tr}\left(L_1^\dagger + L_2^\dagger\right) - 2 \, {\rm Tr}\left(\sqrt{L_1^{\frac{\dagger}{2}} L_2^\dagger L_1^{\frac{\dagger}{2}}}\right).
\end{equation}

The advantage of this distance over more traditional graph distances (eg. $\ell_2$, graph edit distance...) is that it takes into account the importance of an edge to the graph structure. This allows to better capture topological features in the distance metric. Another advantage is that the Wasserstein distance comes with a transport map that allows to transfer signals from one graph to the other. Hence, the mapping of signals over graphs yields
\begin{equation}
T(x) = L_1^{\frac{\dagger}{2}}\Big(L_1^{\frac{\dagger}{2}} L_2^\dagger  L_1^{\frac{\dagger}{2}}\Big)^{\frac{\dagger}{2}}L_1^{\frac{\dagger}{2}} x,
\end{equation}
which represents the signal $x$, originally living on graph $\mathcal{G}_1$, adapted to the structure of graph $\mathcal{G}_2$.

The above Wasserstein distance $\mathcal{W}_2^2$ requires the two graphs to be of the same size. However, we want to compare graphs of different sizes as well, which represents a common setting in practice. Throughout the rest of this work, we will consider two graphs $\mathcal{G}_1 = (V_1, E_1)$ and $\mathcal{G}_2=(V_2,E_2)$, and we arbitrarily pick $\mathcal{G}_1$ as the graph with the smaller number of nodes.

\begin{figure}
\centering
\includegraphics[width=0.3\textwidth]{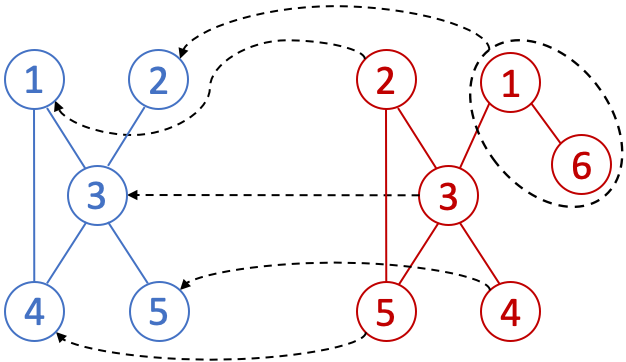}
\caption{One-to-many assignment between different-size graphs.}
\label{fig:example1}
\end{figure}

We now compare graphs of different sizes by looking for the one-to-many assignment between their vertices, similarly to \cite{zaslavskiy2010many}. This is illustrated in the toy example of Figure \ref{fig:example1}, where every vertex of the smaller graph $\mathcal{G}_1$ is assigned to one or more vertices in the larger graph $\mathcal{G}_2$, and every vertex of $\mathcal{G}_2$ is assigned to exactly one vertex in $\mathcal{G}_1$. Let $k_{max}\ge1$ be the maximum number of nodes in $\mathcal{G}_2$ matched to a single node in $\mathcal{G}_1$. Such a one-to-many assignment can be described by a matrix $P\in\mathbb{R}^{|V_1| \times |V_2|}$ satisfying the constraints
\begin{equation}\label{eq:hard_constraint}
\mathcal{C}_{\rm hard} 
=
\left\{
P \in \RR^{|V_1| \times |V_2|}:
\;
\begin{aligned}
&(\forall i, \forall j)\; P_{ij} \in \{0,1\}
\\ 
&(\forall i)\; \textstyle \sum_{j} P_{ij} \in \left[1, k_{\max}\right] 
\\
&(\forall j)\; \textstyle \sum_{i} P_{ij} = 1
\end{aligned}
\right\}
\!.
\end{equation}
In words, the matrix $P$ only takes values zero or one, which corresponds to a hard assignment. Moreover, the sum of each matrix row has to be between $1$ and $k_{\max}$, ensuring that every vertex of $\mathcal{G}_1$ is matched to at least one and at most $k_{\max}$ vertices of $\mathcal{G}_2$. Finally, the sum of each matrix column has to be exactly one, so that every vertex of $\mathcal{G}_2$ is matched to exactly one vertex of $\mathcal{G}_1$. 
To ensure that $\mathcal{C}_{\rm hard}$ is a nonempty constraint set, we require that 
\begin{equation}\label{eq:kmax_condition}
1 \le k_{\max} \le 1 + |V_2| - |V_1|.
\end{equation}

Given the true assignment matrix $P_* \in \mathcal{C}_{\rm hard}$, the larger graph $\mathcal{G}_2$ can be aligned to the smaller graph $\mathcal{G}_1$ by transforming its Laplacian matrix as $P_* L_2 P_*^\top$ \cite{zaslavskiy2010many}, yielding an associated distribution of signals:
\begin{equation}\label{eq:perfect_assignment}
\mu^{\mathcal{G}_2}_{P_*} = \mathcal{N}\big(0, (P_* L_2 P_*^\top)^\dagger\big).
\end{equation}
The graph alignment with the one-to-many assignment solution thus naturally leads to the use of $\mathcal{W}_2^2(\nu^{\mathcal{G}_1}, \mu^{\mathcal{G}_2}_{P_*})$ of Equation (\ref{eq:wass_explicit_graph}) for evaluating the distance\footnote{It is not a distance in the theoretical sense. For brevity, we will use the term ``distance" with an abuse of terminology.} between graphs that originally have different sizes.

Of course, the true assignment matrix  $P_*$ is often unknown beforehand. We are thus interested in estimating the best alignment, or equivalently in finding the assignment matrix $P$ that minimizes the distance between two graphs $\mathcal{G}_1$ and $\mathcal{G}_2$, leading to the optimization problem
\begin{equation}\label{eq:hard_assignment_problem}
\minimize{P \in \mathcal{C}_{\rm hard}} \mathcal{W}_2^2\big(\nu^{\mathcal{G}_1}, \mu^{\mathcal{G}_2}_{P}\big).
\end{equation}
The main difficulty in solving Problem \eqref{eq:hard_assignment_problem} arises from the constraint $\mathcal{C}_{\rm hard}$ defined in \eqref{eq:hard_constraint}, since it leads to a discrete optimization problem with a factorial number of feasible solutions. To circumvent this issue, we propose a relaxation of the one-to-many assignment problem in the next section.

\section{Optimization algorithm}
\label{sec:optimization}

To deal with the nonconvexity of the alignment problem in Equation \eqref{eq:hard_assignment_problem}, we rely on two main ideas. Firstly, we relax the binary constraint into the unitary interval, so that $P$ becomes a \emph{soft}-assignment matrix belonging to the set 
\begin{equation}
\mathcal{C}_{\rm soft}
=
\left\{
P \in \RR^{|V_1| \times |V_2|}:
\;
\begin{aligned}
&(\forall i, \forall j)\; P_{ij} \in [0,1]
\\ 
&(\forall i)\; \textstyle \sum_{j} P_{ij} \in \left[1, k_{\max}\right] 
\\
&(\forall j)\; \textstyle \sum_{i} P_{ij} = 1
\end{aligned}
\right\}
\!.
\end{equation}
Secondly, we enforce the relaxed constraints implicitly using the Dykstra operator
\begin{equation}
\mathcal{A}_{\tau} \colon \RR^{|V_1| \times |V_2|} \to \mathcal{C}_{\rm soft},
\end{equation}
which transforms a rectangular matrix into a soft-assignment matrix, as explained in Section \ref{sec:dykstra}. This operator can be injected into the cost function to remove all the constraints, thus yielding the new unconstrained optimization problem
\begin{equation}\label{eq:relaxed_problem}
\minimize{\widetilde{P}\in \RR^{|V_1| \times |V_2|}}{}\; \mathcal{W}_2^2\Big(\nu^{\mathcal{G}_1}, \mu^{\mathcal{G}_2}_{\mathcal{A}_\tau(\widetilde{P})}\Big).
\end{equation}

Problem \eqref{eq:relaxed_problem} is highly nonconvex, which may cause gradient descent to converge towards a local minimum. As we will see in Section \ref{sec:stochastic}, 
using the Dykstra operator $\mathcal{A}_\tau(\widetilde{P})$ will allow us to devise a stochastic formulation that can be efficiently solved with a variant of gradient descent integrating Bayesian exploration in the optimization process, possibly helping the algorithm to find better local minima.

\subsection{Dykstra operator}\label{sec:dykstra}
Given a rectangular matrix $\widetilde{P}$ and a small constant $\tau>0$, the Dykstra operator normalizes the rows and columns of $\exp(\widetilde{P}/\tau)$ to obtain a one-to-many assignment matrix, where a node in the smaller graph is matched to one or more (but at most $k_{\max}$) nodes in the larger graph. It is defined as
\begin{equation}\label{eq:Dykstra}
\mathcal{A}_\tau(\widetilde{P}) = \operatorname*{argmax}_{P\in\mathcal{C}_{\rm soft}} \left[ \big\langle P, \widetilde{P} \big\rangle - \tau \sum_{ij} P_{ij} \log(P_{ij})\right].
\end{equation}
This operator can be efficiently computed by the Dykstra algorithm \cite{dykstra1983} with Bregman projections \cite{bauschke2000}. Indeed, Problem \eqref{eq:Dykstra} can be written as a Kullback-Leibler (KL) projection \cite{benamou2015}
\begin{equation}\label{eq:kld_projection}
\mathcal{A}_\tau(\widetilde{P}) = \operatorname*{argmin}_{P\in\mathcal{C}^{(0)} \cap \mathcal{C}^{(1)}} {\rm KL}\big(P \,|\, \exp({\widetilde{P}/\tau})\big), 
\end{equation}
with 
\begin{equation}
\begin{aligned}
\mathcal{C}^{(0)} &= 
\big\{ \Xi \in \RR_+^{|V_1| \times |V_2|} \;|\; \Xi \mathbbm{1}_{|V_2|} \in \left[1, k_{\max}\right]^{|V_1|} \big\},
\\
\mathcal{C}^{(1)} &= \big\{ \Xi \in \RR_+^{|V_1| \times |V_2|} \;|\; \Xi^\top \mathbbm{1}_{|V_1|}  = \mathbbm{1}_{|V_2|} \big\}.
\end{aligned}
\end{equation}

The Dykstra algorithm starts by initializing
\begin{equation}
P^{[0]} = \exp({\widetilde{P}/\tau})
\quad{\rm and}\quad
Q^{[0]} = Q^{[-1]} = \mathbbm{1}_{|V_1| \times |V_2|},
\end{equation}
and then iterates for every $t=0,1,\dots$
\begin{align}
P^{[t+1]} &= \mathcal{P}^{\rm KL}_{\mathcal{C}^{(t {\,\rm mod\,} 2)}}\big(P^{[t]} \odot Q^{[t-1]}\big),\\
Q^{[t+1]} &= \frac{Q^{[t-1]} \odot P^{[t]}}{P^{[t+1]}},
\end{align}
where all operations are meant entry-wise.\footnote{$\odot$ denotes the entry-wise (Hadamard) product of matrices.} The KL projections are defined, for every $\Xi \in \RR_+^{|V_1| \times |V_2|}$, as follows
\begin{align}
\mathcal{P}^{\rm KL}_{\mathcal{C}^{(0)}}\big(\Xi\big) &= \operatorname{diag}\left(\left[\frac{\max\big\{1, \min\big\{\sum_j \Xi{ij}, k_{\max}\big\}\big\}}{\sum_j \Xi{ij}}\right]_{i}\right) \Xi \\
\mathcal{P}^{\rm KL}_{\mathcal{C}^{(1)}}\big(\Xi\big) &= 
\Xi\operatorname{diag}\left(\left[\frac{1}{\sum_i \Xi{ij}}\right]_{j}\right).
\end{align}
In the limit $\tau\to0$, the operator $\mathcal{A}_{\tau}$ yields a one-to-many assignment matrix. It is also differentiable \citep{Luise2018}, and can be thus used in a cost function optimized by gradient descent, as we will see in Section \ref{sec:stochastic}.

\subsubsection{Connection to Sinkhorn}\label{sec:sinkhorn}
In the special case where the two graphs have the same size $|V_1|=|V_2|=|V|$, the condition in \eqref{eq:kmax_condition} leads to $k_{\max}=1$, and thus $\mathcal{C}_{\rm soft}$ reduces to the space of doubly-stochastic matrices. The Dykstra operator then reverts to a Sinkhorn operator \citep{Sinkhorn1964, Cuturi2013, Genevay2018a, Mena2018, NIPS2019_9539}. Given a square matrix $\widetilde{P}$ and a small constant $\tau>0$, the Sinkhorn operator normalizes the rows and columns of $\exp(\widetilde{P}/\tau)$ so as to obtain a doubly stochastic matrix. Formally, it is defined as
\begin{equation}\label{eq:sinkhorn}
\mathcal{S}_\tau(\widetilde{P}) = \operatorname*{argmax}_{P\in\mathcal{C}_{\rm doubly}} \left[ \left\langle P, \widetilde{P} \right\rangle - \tau \sum_{ij} P_{ij} \log(P_{ij})\right], 
\end{equation}
where $\mathcal{C}_{\rm doubly}$ is the set of doubly stochastic matrices
\begin{equation}
\mathcal{C}_{\rm doubly} =
\left\{
P \in \RR^{|V| \times |V|}:
\;
\begin{aligned}
&(\forall i, \forall j)\; P_{ij} \in [0,1]
\\ 
&(\forall i)\; \textstyle \sum_{j} P_{ij} = 1 
\\
&(\forall j)\; \textstyle \sum_{i} P_{ij} = 1
\end{aligned}
\right\}
\!.
\end{equation}
It is well known that the above operator can be computed with the following iterations
\begin{equation}
\begin{aligned}
P^{[0]} &= \exp(\widetilde{P}/\tau) \\
L^{[t]} &= \operatorname{diag}\big( P^{[t]} \mathbbm{1}_{|V|} \big)^{-1} \\
R^{[t]} &= \operatorname{diag}\big(\mathbbm{1}^\top_{|V|} L^{[t]} P^{[t]} \big)^{-1} \\[0.5em]
P^{[t+1]} &= L^{[t]} P^{[t]} R^{[t]}.
\end{aligned}
\end{equation}
In the limit $\tau\to0$, the operator $S_{\tau}$ yields a permutation matrix \citep{Mena2018}. It is also differentiable \citep{Luise2018}, and can be thus used in a cost function optimized by gradient descent, as we will see in Section \ref{sec:stochastic}.

\begin{figure*}

    \begin{minipage}[b]{0.99\textwidth}
    \centering
    \includegraphics[width=0.65\textwidth,height=3.5cm]{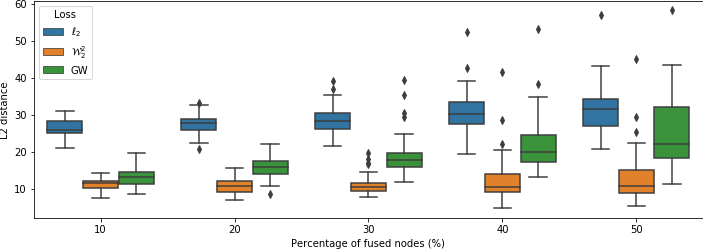}
    \end{minipage}    
    \vfill 
    \begin{minipage}[b]{0.99\textwidth}
    \centering
    \includegraphics[width=0.65\textwidth,height=3.5cm]{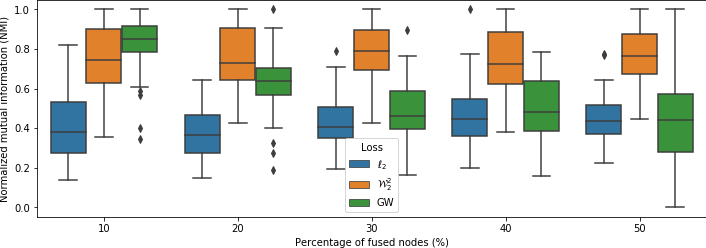}
    \end{minipage}  
    
    \caption{Alignment and detection of communities in structured graphs, showing the recovery of local changes, as well as the global graph structure. The experiment is performed on distorted stochastic block model graphs as a function of the percentage of fused nodes. The graph $\mathcal{G}_2$ is a four stochastic block model with 24 nodes. The graph $\mathcal{G}_1$ is a random distorted version of $\mathcal{G}_2$, where the edges are collapsed until the target percentage of nodes is fused. 
    We compare three different distances: Gromov-Wasserstein (GW), the $\ell_2$ defined as $\|L_1-PL_2P^\top\|^2$ and solved using our stochastic algorithm and the proposed distance $\mathcal{W}_2^2\big(\nu^{\mathcal{G}_1}, \mu^{\mathcal{G}_2}_{P}\big)$.
    The first plot shows the $\ell_2$ distance between aligned graphs (closer to 0 the better), while the second one shows the community detection performance using spectral clustering technique in terms of Normalized Mutual Information (NMI closer to 1 the better).
    \label{fig:community_distorted}}
\end{figure*}

\subsection{Stochastic formulation}\label{sec:stochastic}
With help of the Dykstra operator, the cost function in Problem \eqref{eq:relaxed_problem} becomes differentiable, and can be thus optimized by gradient descent. However, the nonconvex nature of the problem may cause gradient descent to converge towards a local minimum. Instead of directly solving Problem \eqref{eq:relaxed_problem}, we propose to optimize the expectation w.r.t.\ the parameters $\theta$ of some distribution $q_\theta$, yielding
\begin{equation}\label{eq:relaxed_problem_stochastic}
\operatorname*{minimize}_{\theta} \; \mathbb{E}_{\widetilde{P}\sim q_\theta}\Big\{\mathcal{W}_2^2\big(\nu^{\mathcal{G}_1}, \mu^{\mathcal{G}_2}_{\mathcal{A}_\tau(\widetilde{P})}\big) \Big\}.
\end{equation}
The optimization of the expectation w.r.t.\ the parameters $\theta$ aims at shaping the distribution $q_\theta$ so as to put all its mass on a minimizer of the original cost function, thus integrating the use of Bayesian exploration in the optimization process, possibly helping the algorithm to find better local minima.

A standard choice for $q_\theta$ in continuous optimization is the multivariate normal distribution, leading to $\theta=(\eta,\sigma)$ with $\eta$ and $\sigma$ being $|V_1| \times |V_2|$ matrices. By leveraging the reparameterization trick \citep{Kingma2014, Figurnov2018}, which boils down to setting
\begin{equation}
\widetilde{P}_{ij} = \eta_{ij} + \sigma_{ij} \epsilon_{ij}
\qquad\textrm{with}\qquad
\epsilon_{ij} \sim \mathcal{N}(0,1).
\end{equation}
The problem of Equation \eqref{eq:relaxed_problem_stochastic} can thus be reformulated as
\begin{equation}\label{eq:final_problem}
\operatorname*{minimize}_{\eta,\sigma}{}\; \mathcal{J}(\eta, \sigma)  := \mathbb{E}_{\epsilon\sim q_{\sf unit}}
\Big\{ 
\mathcal{W}_2^2\big(\nu^{\mathcal{G}_1}, \mu^{\mathcal{G}_2}_{\mathcal{A}_\tau(\eta + \sigma \odot \epsilon)}\big)
\Big\},
\end{equation}
where $q_{\sf unit} = \prod_{i,j}\mathcal{N}(0,1)$ denotes the multivariate normal distribution with zero mean and unitary variance. 
The advantage of this reformulation is that the gradient of the above stochastic function can be approximated by sampling from the parameterless distribution $q_{\sf unit}$, yielding
\begin{equation}\label{eq:approx_gradient}
\nabla \mathcal{J}(\eta, \sigma) \approx 
\sum_{\epsilon\sim q_{\sf unit}} \nabla
\mathcal{W}_2^2\big(\nu^{\mathcal{G}_1}, \mu^{\mathcal{G}_2}_{\mathcal{A}_\tau(\eta + \sigma \odot \epsilon)}\big).
\end{equation}
The problem can be thus solved by stochastic gradient descent \citep{Khan2017}. Our approach is summarized in Algorithm \ref{algo}. 

Under mild assumptions, the algorithm converges almost surely to a critical point, which is not guaranteed to be the global minimum, as the problem is nonconvex. The computational complexity of a naive implementation is $O(N^3)$ per iteration, due to the matrix square-root operation, but faster options exist to approximate this operation \cite{Lin2017}. Moreover, the computation of pseudo-inverses can be avoided by adding a small diagonal shift to the Laplacian matrices and directly computing the inverse matrices, which is orders of magnitude faster.

\begin{algorithm}[t]
\caption{Approximate solution to Problem \eqref{eq:hard_assignment_problem}.}
\label{algo} 
\begin{algorithmic}[1]
	\STATE{\bf Input:} {Graphs $\mathcal{G}_1$ and $\mathcal{G}_2$} 
	\STATE{\bf Input:} {Sampling $S\in\mathbb{N}$, step size $\gamma>0$, and $\tau>0$}
	\STATE{\bf Input:} {Random initialization of matrices $\eta_0$ and $\sigma_0$}
	
	\FOR{$t=0,1,\dots$}
	\STATE Draw samples $\{\epsilon_t^{(s)}\}_{1\le s\le S}$ from $q_{\sf unit}$
	\STATE Approximate the cost function with Equation \eqref{eq:final_problem} $$ \mathcal{J}_t(\eta_t,\sigma_t) = \frac{1}{S} \sum_{s=1}^S 
	\mathcal{W}_2^2\Big(\nu^{\mathcal{G}_1}, \mu^{\mathcal{G}_2}_{\mathcal{A}_\tau(\eta_t + \sigma_t \odot \epsilon_s)}\Big)$$
	\STATE $g_t \leftarrow $ gradient of $ \mathcal{J}_t$ evaluated at $(\eta_t,\sigma_t)$
	\STATE $(\eta_{t+1},\sigma_{t+1}) \leftarrow $ update of $(\eta_t,\sigma_t)$ with step size $\gamma$ using $g_t$
	\ENDFOR
	\STATE \textbf{Output:} $P = \mathcal{A}_{\tau}(\eta_\infty)$
\end{algorithmic}
\end{algorithm}

\section{Experiments}
\label{sec:experiments}
We now analyse the performance of our new algorithm in two parts. Firstly, we assess the performance achieved by our approach for graph alignment and community detection in structured graphs, testing the preservation of both local and global graph properties. We investigate the influence of distance on alignment recovery and compare to methods using different definitions of graph distance for graph alignment. Secondly, we extend our analysis to graph classification, where we compare our approach with several state-of-the-art methods. 

Prior to running experiments, we determined the algorithmic parameters $\tau$ (in the Dykstra operator) and $\gamma$ (step size in SGD) with grid search, while $S$ (sampling size) was fixed empirically. In all experiments, we set $\tau=3$, $\gamma=1$ and $S=10$. We set the maximal number of Dykstra iterations to 20, and we run stochastic gradient descent for 1000 iterations. As our algorithm seems robust to different initialisations, we used random initialization in all our experiments. The algorithm was implemented in PyTorch with AMSGrad method \cite{j.2018on}.

\begin{figure*}
    
    \begin{center}
    \begin{minipage}[b]{0.99\textwidth}
    \centering
    \includegraphics[width=0.65\textwidth,height=3.5cm]{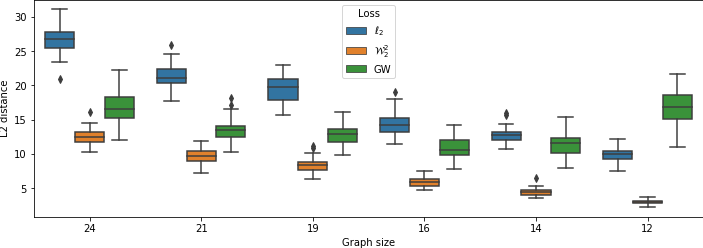}
    \end{minipage}    
    \vfill 
    \begin{minipage}[b]{0.99\textwidth}
    \centering
    \includegraphics[width=0.65\textwidth,height=3.5cm]{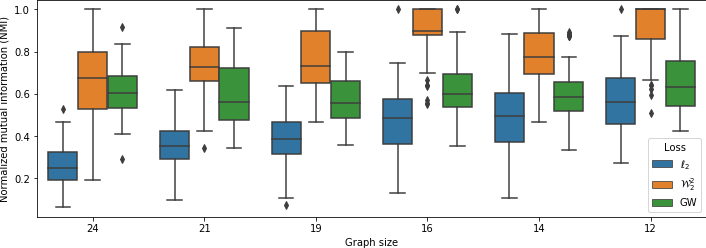}
    \end{minipage}  
    \end{center}

    \caption{Alignment and detection of communities in structured graphs, showing the recovery of local changes, as well as the global graph structure. The experiment is performed on random instances of stochastic block model graphs as a function of the graph size. The graph $\mathcal{G}_2$ is a four stochastic block model with 24 nodes. The graph $\mathcal{G}_1$ is a random graph with four stochastic block model with different number of nodes. 
    We compare three different methods: Gromov-Wasserstein (GW), the $\ell_2$ defined as $\|L_1-PL_2P^\top\|^2$ and solved using our stochastic algorithm and the proposed distance $\mathcal{W}_2^2\big(\nu^{\mathcal{G}_1}, \mu^{\mathcal{G}_2}_{P}\big)$.
    The first plot shows the $\ell_2$ distance between aligned graphs (closer to 0 the better), while the second one shows the community detection performance using spectral clustering technique in terms of Normalized Mutual Information (NMI closer to 1 the better).}
    \label{fig:community_size}
\end{figure*}
\subsection{Graph alignment and community detection}
In this section, we test our proposed approach for graph alignment and recovery of communities in structured graphs. Namely, apart from the direct comparison of two graphs matrices, we evaluate the preservation of global properties by comparing the clustering of nodes into communities. We consider two experimental settings. In the first one (Figure \ref{fig:community_distorted}), we generate a stochastic block model graph $\mathcal{G}_2$ with 24 nodes and 4 communities. The graph $\mathcal{G}_1$ is a noisy version of $\mathcal{G}_2$ constructed by randomly collapsing edges, merging two connected nodes into one, until a target percentage of nodes is merged. We then generate a random permutation to change the order of the nodes in graph $\mathcal{G}_1$. 

In the second experimental setting (Figure \ref{fig:community_size}), the graph $\mathcal{G}_2$ is again generated as a stochastic block model with four communities. For each $\mathcal{G}_2$, six graphs $\mathcal{G}_1$ are created as random instances of stochastic block model graphs with the same number of communities, but with a different number of vertices and edges. Apart from the number of communities, there is no direct connection between $\mathcal{G}_1$ and $\mathcal{G}_2$.

We investigate the influence of a distance metric on alignment recovery. We compare three different methods for graph alignment, namely the proposed method based on the Wasserstein distance between graphs, the proposed stochastic algorithm with the Euclidean distance ($\ell_2$), and the state-of-the-art Gromov-Wasserstein distance \cite{2016-peyre-icml} for graphs (GW), using the Euclidean distance between shortest path matrices, as proposed in \cite{vayer2018optimal}. We repeat each experiment 50 times, after adjusting parameters for all compared methods, and show the results in Figures \ref{fig:community_distorted} and \ref{fig:community_size}.

We now evaluate the structure recovery of the community-based models through spectral clustering. Namely, after alignment estimation, we cluster the nodes in both graphs. A good alignment should detect and preserve communities, keeping the nodes in the same clusters, close to their original neighbours, even when the exact neighbours are not recovered. We evaluate the quality of community recovery with normalized mutual information (NMI) between the clusters in the original graph and the recovered clusters. We further evaluate the alignment quality by checking the difference between the two graphs in terms of the $\ell_2$ norm. While it is not the best possible distance measure for graphs, it is used here as a complementary measure to the NMI, not taking any special structural information into account. It can also be seen as an unbiased metric to compare the two methods performing the best in terms of NMI.

As shown in Figure \ref{fig:community_distorted}, the proposed approach manages to capture the structural information and outperform methods based on different distance metrics, especially under large perturbations. 
In Figure \ref{fig:community_size}, we observe an increase in performance in terms of NMI for both $\ell_2$ and $\mathcal{W}_2^2$. The emergence of this phenomenon despite the growing size difference between compared graphs suggests our assignment matrix has the ability to fuse nodes into meaningful groups, forming well defined clusters.

\subsection{Graph classification}

We now tackle the task of graph classification on two different datasets: PTC \cite{NIPS2016_6166} and IMDB-B \cite{deep_graph_kernels2015}. We randomnly sample 100 graphs from each dataset. The graphs have a different number of nodes and edges. We use $\mathcal{W}_2^2$ to align graphs and compute graph distances, and eventually use a simple non-parametric 1-NN classification algorithm to classify
graphs. We compare the classification performance with methods where the same 1-NN classifier is used with different state-of-the-art methods for graph alignment: GW \cite{2016-peyre-icml, vayer2018optimal}, GA \cite{gold1996graduated}, IPFP \cite{leordeanu2009integer}, RRWM \cite{cho2010reweighted}, NetLSD \cite{tsitsulin2018netlsd}, and the proposed stochastic algorithm with the Euclidean distance ($\ell2$) instead of the Wasserstein distance in Eq.~\eqref{eq:relaxed_problem_stochastic} . We present the accuracy scores in Table \ref{tab:ptc}, where the classification with the proposed $\mathcal{W}_2^2$ clearly outperforms the other methods in terms of general accuracy. Furthermore, we analyse the performance of $\mathcal{W}_2^2$, GW and $\ell_2$ on several examples from the two datasets.

\begin{table*}
  \centering
    \caption{Accuracy scores for 1-NN classification results on graph dataset.}
    \label{tab:ptc}
    \small
    \begin{tabular}{c | c c c c c c c} 
    \toprule
        Dataset & GA    & IPFP  & RRWM  & GW    & NetLSD & $\ell_2$ & $\mathcal{W}_2^2$ \\ 
      \midrule
       IMDB-B & 56.72 & 55.22 & 61.19 & 54.54  & 53.73 & 54.54 & \textbf{63.63} \\
       PTC    & 50.75 & 52.24 & 49.25 & 56.71  & 52.23 & 47.76 & \textbf{61.19}  \\
      \bottomrule
    \end{tabular}
\end{table*}

\newcommand{\filler}{\hspace{1.5cm}}

\begin{figure}

    \begin{minipage}[t]{0.5\textwidth}
    \centering
    {\footnotesize
    \begin{tabular}{l c c r}
    $\mathcal{G}_1$\filler & \filler$\mathcal{G}_2$\filler & \filler$\mathcal{G}_3$\filler & \filler$\mathcal{G}_4$\filler
        \end{tabular}}
    \end{minipage}

    \begin{minipage}[b]{0.5\textwidth}
    \includegraphics[width=0.99\textwidth]{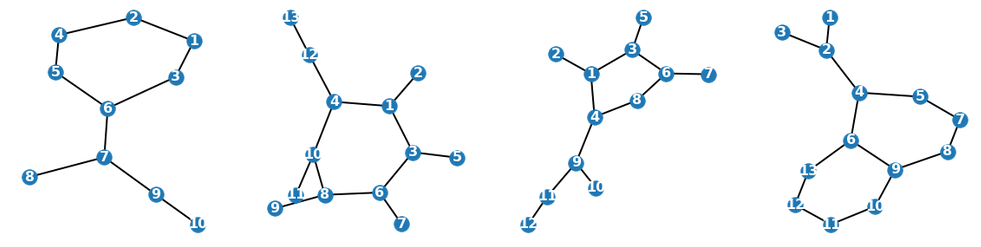}
    \end{minipage}
    
    \vfill
    
    \begin{minipage}[t]{0.5\textwidth}
    \centering
    {\footnotesize
    \begin{tabular}{l c c c c c c}
    \toprule
        & & $\mathcal{D}(\mathcal{G}_1,\mathcal{G}_2)$ & $<$ & $\mathcal{D}(\mathcal{G}_1,\mathcal{G}_3)$ & $>$ & $\mathcal{D}(\mathcal{G}_3,\mathcal{G}_4)$\\ \cmidrule{3-7}
         $\ell_2$-norm       && 0.0058 && 0.0096 && 0.0093 \\	
         GW                 && 1.2417 && 0.7866 && 2.2204 \\ 
         $\mathcal{W}_2^2$  && \textbf{0.9301} && \textbf{0.9465} && \textbf{0.5457} \\ 
         \bottomrule
        \end{tabular}}
    \end{minipage}
    
    \vspace{1em}
    
    \begin{minipage}[b]{0.5\textwidth}
    \includegraphics[width=0.99\textwidth]{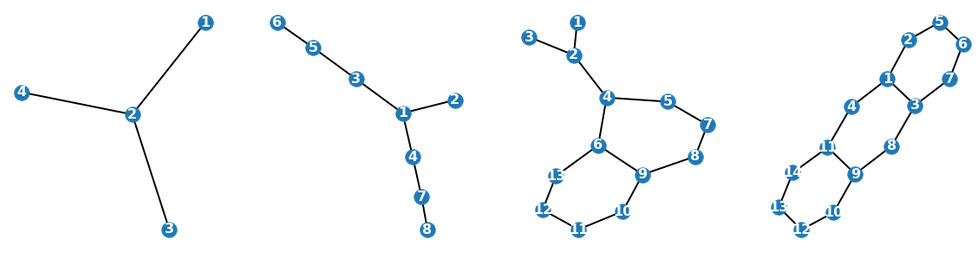}
    \end{minipage}
    
    \vfill
    
    \begin{minipage}[t]{0.5\textwidth}
    \centering
    {\footnotesize
    \begin{tabular}{l c c c c c c}
        \toprule
        & & $\mathcal{D}(\mathcal{G}_1,\mathcal{G}_2)$ & $<$& $\mathcal{D}(\mathcal{G}_1,\mathcal{G}_3)$ & $>$& $\mathcal{D}(\mathcal{G}_3,\mathcal{G}_4)$\\ \cmidrule{3-7}
         $\ell_2$-norm      && 0.0476 && 0.0002 && 0.0067 \\	
         GW                  && \textbf{2.7187} && \textbf{3.5081}&& \textbf{0.9897} \\ 
         $\mathcal{W}_2^2$   && \textbf{1.0891} && \textbf{2.0754}&& \textbf{0.8202} \\ 
         \bottomrule
        \end{tabular}}
    \end{minipage}
    
    \vspace{1em}
    
    \begin{minipage}[b]{0.5\textwidth}
    \includegraphics[width=0.99\textwidth]{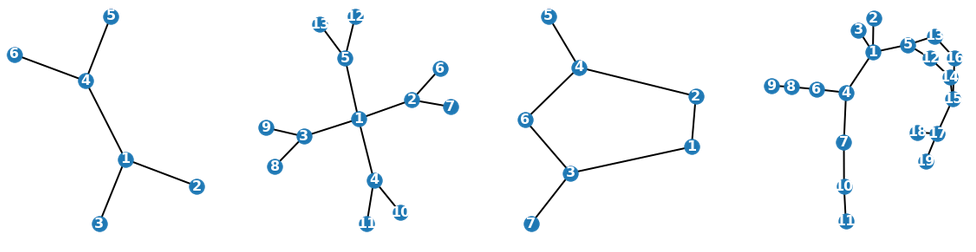}
    \end{minipage}
    
    \begin{minipage}[t]{0.5\textwidth}
    \centering
    {\footnotesize
    \begin{tabular}{l c c c c c c}
        \toprule
        & & $\mathcal{D}(\mathcal{G}_1,\mathcal{G}_2)$ & $<$& $\mathcal{D}(\mathcal{G}_1,\mathcal{G}_3)$ &$>$ & $\mathcal{D}(\mathcal{G}_3,\mathcal{G}_4)$\\ \cmidrule{3-7}
         $\ell_2$-norm       && 0.1580 && 0.0023 && 0.0050 \\	
         GW                 && 1.4493 && 0.9217 && 8.5444 \\ 
         $\mathcal{W}_2^2$   && 1.2069 && 0.2332&& 1.7364 \\ 
         \bottomrule
        \end{tabular}}
    \end{minipage}
    
    \caption{PTC dataset with two classes. Each row presents a set of graph examples, from the left to the right: $\mathcal{G}_1$,
$\mathcal{G}_2$, $\mathcal{G}_3$ and $\mathcal{G}_4$. 
$\mathcal{G}_1$ and $\mathcal{G}_2$ belong to class 0.
$\mathcal{G}_3$ and $\mathcal{G}_4$ belong to class 1. 
Each table provides two kind of distances: an intra ($\mathcal{D}(\mathcal{G}_1,\mathcal{G}_2)$ and $\mathcal{D}(\mathcal{G}_3,\mathcal{G}_4)$) and inter ($\mathcal{D}(\mathcal{G}_1,\mathcal{G}_3)$) classes. We evaluate three different methods in terms of distances in order to classify the graphs \big(e.g. $\mathcal{D}(\mathcal{G}_1,\mathcal{G}_2)\le \mathcal{D}(\mathcal{G}_1,\mathcal{G}_3)$ or $\mathcal{D}(\mathcal{G}_3,\mathcal{G}_4)\le\mathcal{D}(\mathcal{G}_1,\mathcal{G}_3)$\big).
\label{fig:ptc}} 
\end{figure}

\begin{figure}

    \begin{minipage}[t]{0.5\textwidth}
    \centering
    {\footnotesize
    \begin{tabular}{l c c r}
    $\mathcal{G}_1$\filler & \filler$\mathcal{G}_2$\filler & \filler$\mathcal{G}_3$\filler & \filler$\mathcal{G}_4$\filler
        \end{tabular}}
    \end{minipage}
    
    \begin{minipage}[b]{0.5\textwidth}
    \includegraphics[width=0.99\textwidth]{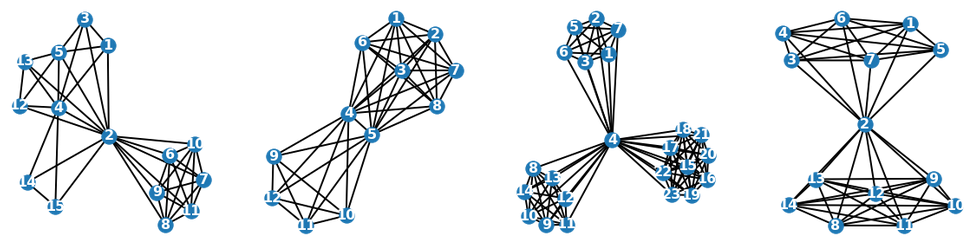}
    \end{minipage}
    
    \begin{minipage}[t]{0.5\textwidth}
    \centering
    {\footnotesize
    \begin{tabular}{l c c c c c c}
    \toprule
        & & $\mathcal{D}(\mathcal{G}_1,\mathcal{G}_2)$ & $<$& $\mathcal{D}(\mathcal{G}_1,\mathcal{G}_3)$ & $>$& $\mathcal{D}(\mathcal{G}_3,\mathcal{G}_4)$\\ \cmidrule{3-7}
         $\ell_2$-norm      && \textbf{0.0083} && \textbf{7.0609} && 8.7336 \\	
         GW               &&  0.3166 && 0.1755 && 0.3096 \\ 
         $\mathcal{W}_2^2$  && \textbf{0.5251} && \textbf{0.6327} && 0.7653 \\ 
         \bottomrule
        \end{tabular}}
    \end{minipage}
    
    \vspace{1em}
    
    \begin{minipage}[b]{0.5\textwidth}
    \includegraphics[width=0.99\textwidth]{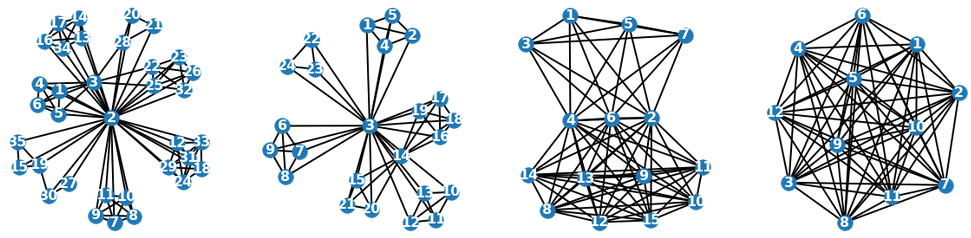}
    \end{minipage}
    
    \vfill
    
    \begin{minipage}[t]{0.5\textwidth}
    \centering
    {\footnotesize
    \begin{tabular}{l c c c c c c}
    \toprule
        & & $\mathcal{D}(\mathcal{G}_1,\mathcal{G}_2)$ & $<$& $\mathcal{D}(\mathcal{G}_1,\mathcal{G}_3)$ & $>$& $\mathcal{D}(\mathcal{G}_3,\mathcal{G}_4)$\\ \cmidrule{3-7}
         $\ell_2$-norm      && 15.1141 && 0.0859 && 0.0084 \\	
         GW                && \textbf{0.1362} && \textbf{0.6224}&& \textbf{0.3233} \\ 
         $\mathcal{W}_2^2$  && \textbf{0.7313} && \textbf{1.4359} && \textbf{0.3120} \\ 
          \bottomrule
        \end{tabular}}
    \end{minipage}
  
   \vspace{1em}
    
    \begin{minipage}[b]{0.5\textwidth}
    \includegraphics[width=0.99\textwidth]{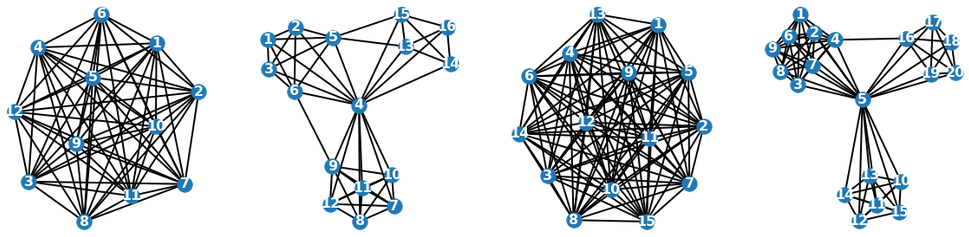}
    \end{minipage}
    
    \vfill
    
    \begin{minipage}[t]{0.5\textwidth}
    \centering
    {\footnotesize
    \begin{tabular}{l c c c c c c}
    \toprule
        & & $\mathcal{D}(\mathcal{G}_1,\mathcal{G}_2)$ & $<$& $\mathcal{D}(\mathcal{G}_1,\mathcal{G}_3)$ & $>$& $\mathcal{D}(\mathcal{G}_3,\mathcal{G}_4)$\\ \cmidrule{3-7}
         $\ell_2$-norm      && 8.7374 && 4.5367 && 1.2624\\	
         GW                 && 0.5998 && 0.0388 && 0.6294 \\ 
         $\mathcal{W}_2^2$  && 0.5529 && 0.3003 && 0.6718 \\ 
          \bottomrule
        \end{tabular}}
    \end{minipage}
    \caption{IMDB-B dataset with two classes. Each row presents a set of graph examples, from the left to the right: $\mathcal{G}_1$,
$\mathcal{G}_2$, $\mathcal{G}_3$ and $\mathcal{G}_4$. 
$\mathcal{G}_1$ and $\mathcal{G}_2$ belong to class 0.
$\mathcal{G}_3$ and $\mathcal{G}_4$ belong to class 1. 
Each table provides two kind of distances: an intra ($\mathcal{D}(\mathcal{G}_1,\mathcal{G}_2)$ and $\mathcal{D}(\mathcal{G}_3,\mathcal{G}_4)$) and inter ($\mathcal{D}(\mathcal{G}_1,\mathcal{G}_3)$) classes. We evaluate three different methods in terms of distances in order to classify the graphs \big(e.g. $\mathcal{D}(\mathcal{G}_1,\mathcal{G}_2)\le \mathcal{D}(\mathcal{G}_1,\mathcal{G}_3)$ or $\mathcal{D}(\mathcal{G}_3,\mathcal{G}_4)\le\mathcal{D}(\mathcal{G}_1,\mathcal{G}_3)$\big). \label{fig:imdbb}} 
\end{figure}

\subsection*{PTC dataset} PTC dataset contains the molecular structure of the NTP dataset. Figure \ref{fig:ptc} presents a set of graph examples from two different classes (0 and 1). 
In the first example (first row), $\mathcal{W}_2^2$ outperforms both $\ell_2$ and $GW$ in separating the two classes. The distinguishing feature between $\mathcal{G}_1$ and $\mathcal{G}_3$ is the number of nodes that forms the ring, which has been captured by $\mathcal{W}_2^2$, thanks to the soft permutation applied to the larger graph $\mathcal{G}_3$ ($|V_3|>|V_2|$).

The second example shows in a very intuitive way how $\mathcal{W}_2^2$ and GW are able to capture structural similarities in graphs, even when those largely vary in size. This is especially clear when comparing the almost two times larger $\mathcal{W}_2^2(\mathcal{G}_1$, $\mathcal{G}_3)$ and $\mathcal{W}_2^2(\mathcal{G}_1$, $\mathcal{G}_2)$, with structurally very similar $\mathcal{G}_1$ and $\mathcal{G}_2$, and an easy-to-imagine assignment of one node in the graph $\mathcal{G}_1$ to several nodes in the graph $\mathcal{G}_2$. However, it is not always as simple to understand the similarities. The third row shows an example in which all the three methods fail to find structural similarities with graphs in the same class.

\subsection*{IMDB-B dataset} IMDB-B dataset contains two classes: Comedy and science-fiction movies, with several examples shown in Figure \ref{fig:imdbb}. The striking difference between example 2 and 3 shows that, while taking into account the global graph structure can be crucial in distinguishing some samples (second row), it remains a challenging dataset with very similar graphs often belonging to different clusters (third row). This possibly explains the low accuracy across all examined methods. However, example 1 shows the high flexibility of the assignment matrix proposed in our algorithm, where the one-to-many assignment is able to detect that graph $\mathcal{G}_1$ is very close to a graph with 2 communities, even if it technically has 3. This combination of putting emphasis on structural information, and allowing for flexibility might be the reason why  $\mathcal{W}_2^2$ still manages to outperform the other investigated methods.

\section{Conclusion}
\label{sec:conclusion}
In this paper, we have proposed a new method to align graphs of different sizes. Equipped with an optimal transport based approach to compute the distance between two smooth graph distributions associated to each graph, we have formulated a new one-to-many alignment problem to find a soft assignment matrix that minimizes the ``mass" transportation from a fixed distribution to a permuted and partially merged distribution. The resulting nonconvex optimization problem is solved efficiently with a novel stochastic gradient descent algorithm. It allows us to align and compare graphs, and it outputs a structurally meaningful distance. We have shown the performance of the proposed method in the context of graph alignment and graph classification. Our results show that the proposed algorithm outperforms state-of-the-art alignment methods for structured graphs. 

\bibliography{ICML_SGOT}
\bibliographystyle{icml2020}

\end{document}